\documentclass[sigconf]{acmart}

\usepackage{amsmath}
\usepackage{booktabs}
\usepackage{tabularx}
\usepackage{multirow}
\usepackage{microtype}

\newcolumntype{L}[1]{>{\raggedright\arraybackslash}m{#1}}
\newcolumntype{C}[1]{>{\centering\arraybackslash}m{#1}}
\newcolumntype{R}[1]{>{\raggedleft\arraybackslash}m{#1}}
\newcommand{\paperTableSetup}{%
  \renewcommand{\arraystretch}{1.14}%
  \setlength{\tabcolsep}{3pt}}
\newcommand{\paperTableHead}[1]{\textbf{#1}}

\setcopyright{none}
\copyrightyear{}

\acmYear{2027}
\acmConference[Preprint]{Preprint}{xxx}{xxx}
\acmDOI{}
\acmISBN{}
\newcommand{\tracespec}{TraceSpec}
\newcommand{\corpus}{TraceSpec-Traj}

\newcommand{\TV}{\operatorname{TV}}
\newcommand{\clip}{\operatorname{clip}}

\begin{document}
\title{TraceSpec: Understanding and Improving Speculative Decoding for LLM Agents}

\author{Jihao Zhao}
\affiliation{%
  \department{School of Information}
  \institution{Renmin University of China}
  \city{Beijing}
  \country{China}
}

\author{Shuaishuai Zu}
\affiliation{%
  \department{School of Information}
  \institution{Renmin University of China}
  \city{Beijing}
  \country{China}
}

\author{Pengfei Li}
\affiliation{%
  \institution{Huawei Technologies Co., Ltd.}
  \city{Beijing}
  \country{China}
}

\author{Biao Qin}
\authornote{Corresponding authors.}
\affiliation{%
  \department{School of Information}
  \institution{Renmin University of China}
  \city{Beijing}
  \country{China}
}

\author{Guanghuan Fang}
\author{Ben Feng}
\affiliation{%
  \institution{Huawei Technologies Co., Ltd.}
  \city{Hangzhou}
  \country{China}
}

\author{Baochuan Yang}
\authornotemark[1]
\affiliation{%
  \institution{Huawei Technologies Co., Ltd.}
  \city{Hangzhou}
  \country{China}
}

\renewcommand{\shortauthors}{Zhao et al.}

\begin{abstract}
LLM agents execute tasks through repeated model calls over evolving histories of reasoning, actions, and tool feedback. Aggregate speculative-decoding metrics merge these distinct decisions, providing limited guidance on where draft supervision is needed. We present TraceSpec, a framework connecting trajectory evaluation with adaptive draft training. It combines dynamic execution with fixed-prefix replay and introduces TraceSpec-Traj, a corpus of 10,000 training states and 2,500 test states spanning browsing and software engineering. Analysis of five speculative baselines reveals that aggregate acceptance can rise with context length while declining within matched late browsing actions. Workload and phase further distinguish these trends, motivating supervision guided by trajectory conditions. TraceSpec trains on intermediate assistant turns and allocates a fixed prediction budget using state-conditioned estimates of prefix-aware acceptance gaps. Across three target sizes and two temperatures, it improves replay throughput by 12.5\% to 22.6\% over its DSpark base. Qwen3-4B ablations show that adaptive allocation increases accepted length by 3.3\% to 10.0\% over fixed full-turn supervision, with additional comparisons substantiating the roles of state conditioning and consecutive-token agreement. The resulting evaluation resources, trajectory findings, and training strategy connect the analysis of agent execution to practical draft improvement. Our code is available at \url{https://github.com/Robot2050/TraceSpec}.

\end{abstract}

\maketitle

\begin{figure*}[t]
  \centering
  \includegraphics[width=1.0\textwidth]{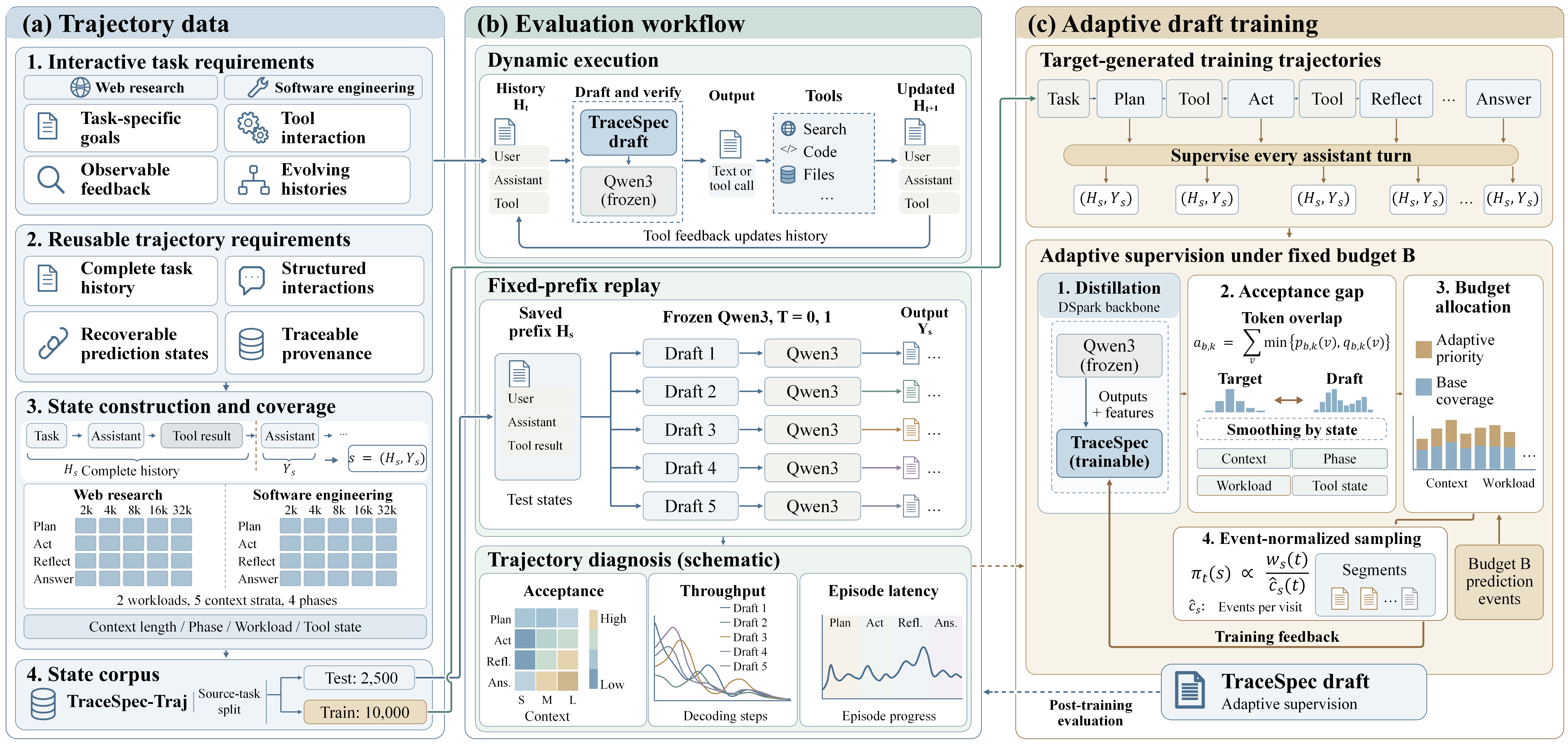}
  \caption{TraceSpec overview. (a) Trajectory-state construction, coverage, and source-task splits. (b) Dynamic execution and fixed-prefix replay with trajectory diagnosis. (c) Full-turn supervision with state-conditioned allocation and event-normalized sampling under a fixed prediction budget.}
  \Description{Three panels connect trajectory data, evaluation, and adaptive draft training. Panel a constructs history and continuation pairs across two workloads, five context strata, and four phases, with source tasks split into training and test partitions. Panel b shows dynamic draft verification by frozen Qwen3, tool feedback updating the history, and five baseline drafters evaluated on identical saved test prefixes. Trajectory diagnostics summarize acceptance, throughput, and episode latency. Panel c expands target-generated training trajectories into assistant segments. Frozen Qwen3 outputs and features supervise a trainable DSpark-based draft. Token overlap contributes to acceptance gaps smoothed by context, phase, workload, and tool state. Base coverage and adaptive priority determine supervision shares, converted to segment sampling probabilities using expected prediction events per visit. Training feedback updates allocation, and the resulting draft returns to evaluation.}
  \label{fig:framework}
\end{figure*}

\section{Introduction}
\label{sec:intro}

Interactive agents use language models to choose actions from information gathered during execution. Web research agents search and inspect documents~\cite{webvoyager}, while software engineering agents navigate repositories and respond to execution feedback~\cite{sweagent}. They interleave reasoning with actions~\cite{react}, learn to invoke tools~\cite{toolformer}, and revise subsequent attempts using textual feedback~\cite{reflexion}. Each step adds information to the history and changes what the next model call must predict. Interactive benchmarks cover shopping~\cite{webshop}, website operation~\cite{webarena}, action generalization across websites~\cite{mind2web}, and tasks spanning diverse environments~\cite{agentbench}. Across these settings, a single task can require many model invocations over progressively longer histories. This repeated interaction makes the distribution of calls within a task central to both serving efficiency and the data needed to adapt a draft model.

Speculative decoding offers a direct way to accelerate these invocations. A small draft model proposes tokens for parallel verification by the target~\cite{leviathan2023}, with acceptance and correction preserving the target sampling distribution~\cite{chen2023}. Accepted length measures how much output each verification round produces. Serving cost also depends on iteration-level scheduling~\cite{orca}, KV-cache management~\cite{vllm}, prefill scheduling~\cite{sarathi}, and attention memory traffic~\cite{flashattention}. Task-level scores describe overall agent capability; understanding speculative efficiency also requires tracing how verification work is distributed among model calls. Relating those calls to their surrounding history processing and tool use clarifies where improved agreement can affect execution time.

Growing histories make context length a natural candidate for explaining drafting difficulty. Long-context research identifies sensitivity to evidence position~\cite{lostmiddle}, variation across tasks in LongBench~\cite{longbench}, and interactions between length and task complexity in RULER~\cite{ruler}. Agent histories additionally encode changing execution states. A long prefix may precede a short tool request, a reflection on new evidence, or a final answer. The same context range can contain decisions from different phases and from different points in a single task. These varying conditions motivate examining acceptance jointly with workload, execution phase, and task progress. The resulting research problem is to characterize agreement across these states and use the observations to allocate a finite supervision budget. A final assistant response covers one decision in a trajectory; earlier actions and revisions provide additional prediction targets. Their value for adaptation depends on where the current draft disagrees with the target, giving a basis for choosing both the decisions to supervise and their training exposure.

We develop \tracespec{} around the decision boundaries in Figure~\ref{fig:framework}, where agents request their next continuations. Dynamic execution on BrowseComp-Plus and SWE-bench Verified captures the model calls encountered during interaction. Fixed-prefix replay on \corpus{} compares drafters on the same saved histories. Together, these protocols examine five speculative baselines across target sizes and sampling temperatures, and resolve their behavior by context, phase, and task progress. The online study shows rising aggregate acceptance alongside falling throughput between the shortest and longest contexts, yet matched late browsing actions exhibit lower acceptance. Replay additionally reveals opposing context trends across workloads and distinct phase profiles. Interpreting these patterns requires knowing what the model is being asked to do at each call, as well as how much history precedes it.

These findings shape two training choices: which assistant turns to supervise and how much exposure to assign them. \tracespec{} uses intermediate decisions from target-generated trajectories and updates their priorities as draft-target disagreement changes. State-conditioned estimates account for trajectory differences, while prefix products capture how early disagreement limits consecutive acceptance. The allocation preserves broad coverage within a common prediction budget and retains the DSpark architecture and distillation objective. Controlled experiments examine supervision scope, adaptive allocation, and the construction of its scoring signal. Section~\ref{sec:framework} develops the framework from trajectory construction through exploratory analysis to draft supervision; Section~\ref{sec:results} presents the comparative results and training ablations. Our contributions are:
\begin{itemize}
  \item We introduce \tracespec{}, an evaluation and supervision framework linking dynamic agent execution to fixed-prefix replay, and \corpus{}, with 10,000 training states and 2,500 test states from disjoint source tasks. Shared decision histories and trajectory annotations support comparisons across workloads, phases, and context strata, making intermediate decisions reusable units for both performance diagnosis and draft adaptation.
  \item We characterize how acceptance profiles vary with workload, execution phase, and drafter. The divergence between aggregate browsing trends and matched late-action behavior shows why the conditions of comparison matter when interpreting long-context performance.
  \item We formulate adaptive draft supervision under a finite prediction budget. Full-turn expansion exposes intermediate decisions, state-conditioned acceptance gaps guide their priority, and event-normalized sampling converts supervision shares into visits while accounting for unequal prediction counts.
  \item We evaluate TraceSpec and five speculative baselines across 108 configurations, relating acceptance to temperature sensitivity and inference cost. TraceSpec leads all 24 replay phase comparisons and increases generation throughput over DSpark in 15 of 18 settings. Four Qwen3-4B ablations examine supervision scope, budget allocation, state conditioning, and consecutive-acceptance scoring.
\end{itemize}

\section{Related Work}
\label{sec:related}

\paragraph{Speculative decoding for agent workloads.}
Drafter designs exploit fused target features in EAGLE-3~\cite{eagle3}, parallel block generation in DFlash~\cite{dflash}, and intra-block dependencies in DSpark~\cite{dspark}. LongSpark uses fixed-size target-state views~\cite{longspark}, while DLite removes the historical draft cache~\cite{dlite}. Agent-oriented approaches further adapt token allocation to batched workflows in AgentSpec~\cite{agentspec} and learn action proposals in LEAP~\cite{leap}. These advances improve proposal construction and execution. Benchmark-level acceptance, however, compresses heterogeneous decisions into a single statistic, obscuring whether a context trend reflects changing agreement within a decision type or a different mixture of calls. Resolving this ambiguity calls for evaluation resources that retain intermediate decisions and permit comparisons under aligned histories.

\paragraph{Draft adaptation and supervision allocation.}
Knowledge distillation provides a basis for draft alignment~\cite{hinton}. DistillSpec examines training data and divergence choices~\cite{distillspec}, online speculative decoding adapts to deployment queries~\cite{osd}, and EDR weights rejection costs by decoding-state occupancy~\cite{edr}. More broadly, curricula organize learning by difficulty~\cite{curriculum}, self-paced learning selects examples using the current model~\cite{selfpaced}, and prioritized replay revisits transitions according to temporal-difference errors~\cite{per}. Importance sampling targets gradient variance~\cite{importance}, while DoReMi adjusts domain mixtures~\cite{doremi}. Applying these ideas to agent trajectories requires attention to both the unit of selection and its contribution to training. A short tool request and an extended reasoning response can supply different numbers of prediction targets even when sampled equally. Deciding which interactions merit further training is therefore coupled with measuring how a limited supervision budget is distributed among them.

\section{TraceSpec Framework}
\label{sec:framework}
The decision record in Figure~\ref{fig:framework} connects measurement and training. The corpus stores the history available before an assistant continuation. The analysis examines acceptance over the encountered workload and within more specific trajectory conditions. Its findings motivate the choice of supervision candidates and the feedback used to assign their training exposure.

\subsection{Trajectory Corpus}
\label{sec:corpus}
\begin{table}[t]
\caption{Trajectory corpus statistics. Source tasks are disjoint across splits.}
\label{tab:corpus}
\centering
\fontsize{8}{10}\selectfont
\paperTableSetup
\begin{tabular*}{\linewidth}{@{}l@{\extracolsep{\fill}\hspace{12pt}}R{38pt}@{\extracolsep{0pt}\hspace{12pt}}R{38pt}@{}}
\toprule
\paperTableHead{Corpus property} & \paperTableHead{Training} & \paperTableHead{Test} \\
\midrule
Trajectory states & 10,000 & 2,500 \\
Source tasks & 5,795 & 1,380 \\
Web research states & 5,000 & 1,250 \\
Software engineering states & 5,000 & 1,250 \\
\midrule
Plan & 2,098 & 501 \\
Act & 3,151 & 920 \\
Reflect & 2,949 & 683 \\
Answer & 1,802 & 396 \\
\bottomrule\end{tabular*}
\end{table}

\tracespec{} represents an agent decision by its available history and next assistant continuation. Each state retains its workload, execution phase, actual tokenized prefix length, preceding tool interactions, and trajectory position. The four phases are planning, action, reflection, and answer generation. The representation in Figure~\ref{fig:framework}(a) supports controlled replay and supervision of intermediate decisions. We construct \corpus{} from DeepResearch-traj~\cite{ipfdata}, Nebius SWE-agent trajectories~\cite{nebiusdata}, and the R2E-Gym development~\cite{r2edata} and testing-agent collections~\cite{r2etestdata}. R2E-Gym supplies repository environments for collecting software engineering trajectories~\cite{r2egym}. Table~\ref{tab:corpus} summarizes the 10,000 training states and 2,500 test states, equally divided between Web research and software engineering. All states from a source task remain in one partition.

Both splits cover all 40 combinations of two domains, five nominal context strata, and four phases. Histories retain preceding assistant decisions and tool observations, making the conditions of each prediction available for analysis. Replay holds this information constant across drafters. Draft adaptation in Section~\ref{sec:method} applies the same representation to target-generated training trajectories, making each assistant continuation a supervision candidate. Joint coverage supports comparisons across contexts and roles; source composition and context quotas appear in Appendix~\ref{app:corpus}.

\begin{figure}[t]
  \centering
  \includegraphics[width=0.47\textwidth]{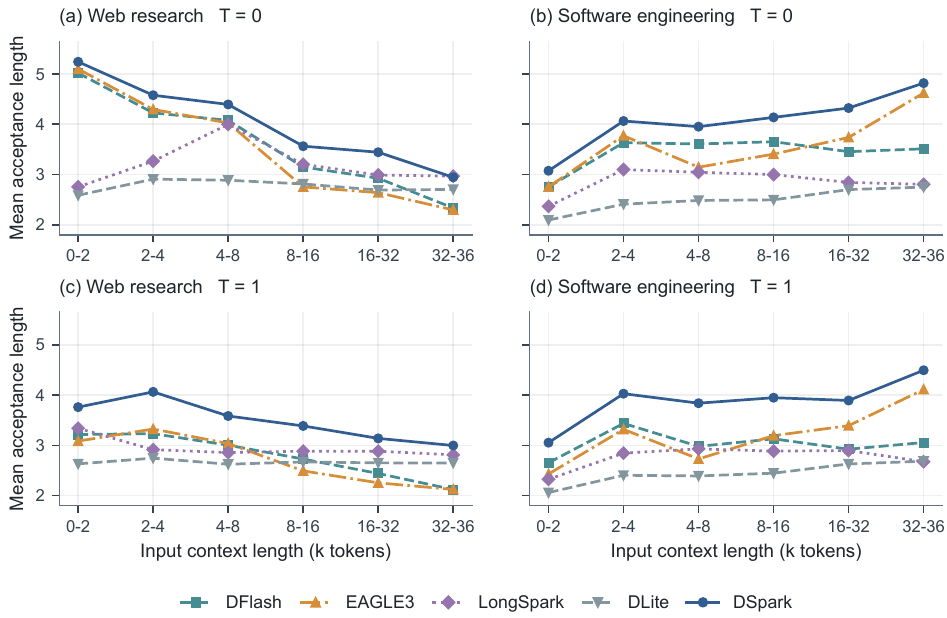}
  \caption{Baseline replay acceptance by context length. Qwen3-8B on \corpus{}, grouped by domain, temperature, and actual input length. All methods receive the same prefixes.}
  \Description{Four line charts show accepted length across context intervals for five draft methods in Web research and software engineering at temperatures zero and one.}
  \label{fig:replaycontext}
\end{figure}

\subsection{Exploratory Trajectory Analysis}
\label{sec:diagnosis}

We investigate how context growth relates to speculative acceptance during agent execution. The analysis follows three levels of evidence: aggregate online behavior, task-paired comparisons within late actions, and workload and phase profiles on fixed replay prefixes. This progression identifies which aspects of the trajectory are relevant to draft supervision.

\begin{figure*}[t]
  \centering
  \includegraphics[width=0.97\textwidth]{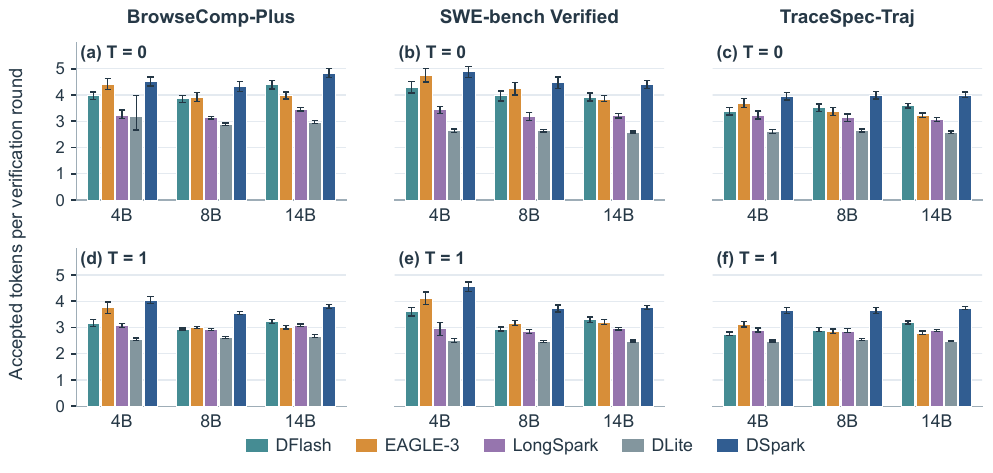}
  \caption{Baseline acceptance across workloads. Columns show dynamic execution on BrowseComp-Plus and SWE-bench Verified, followed by fixed-prefix replay on \corpus{}. Groups denote Qwen3-4B, 8B, and 14B; rows denote $T=0$ and $T=1$. Error bars show pointwise 95\% bootstrap intervals.}
  \Description{Six grouped bar charts in two rows and three columns compare five speculative baselines across three workloads, two temperatures, and three target sizes. Among these baselines, DSpark has the highest overall accepted length in all 18 settings. Accepted length decreases from temperature zero to one for every matched baseline and target.}
  \label{fig:acceptanceoverview}
\end{figure*}

\subsubsection{Analysis Setup}
\label{sec:analysissetup}
Figure~\ref{fig:framework}(b) connects dynamic execution and fixed-prefix replay to trajectory diagnosis. Dynamic execution follows generated actions and tool feedback on BrowseComp-Plus and SWE-bench Verified, while replay supplies identical saved test histories from \corpus{} to each drafter. The former characterizes the workload encountered during interaction; the latter compares methods at common decision boundaries. We measure accepted tokens per verification round, including the correction or bonus token, and model-call throughput, including prefill. Context intervals use $k=1,024$ tokens. The online context analysis fixes Qwen3-8B at $T=0$ and compares DFlash, EAGLE-3, and DSpark; the greedy episode-cost comparison also includes 4B and 14B targets. Replay profiles cover the five baselines with the 8B target at $T=0$ and $T=1$. We progressively condition the analysis on task identity, execution phase, and task progress, then examine workload-specific profiles. Figure~\ref{fig:acceptanceoverview} provides their overall measurements; Section~\ref{sec:setup} specifies the full experimental and statistical protocol.

\subsubsection{Context Growth and Execution Cost}
\label{sec:cost}
\begin{table}[t]
\caption{Mean episode latency (s), $T=0$. BC: BrowseComp-Plus; SWE: SWE-bench Verified. Gain is the latency reduction of DSpark over DFlash.}
\label{tab:latency}
\centering
\fontsize{8}{10}\selectfont
\paperTableSetup
\begin{tabular*}{\linewidth}{@{\extracolsep{\fill}}lccccc@{}}
\toprule
\paperTableHead{Workload} & \paperTableHead{Target} & \paperTableHead{DFlash} & \paperTableHead{EAGLE-3} & \paperTableHead{DSpark} & \paperTableHead{Gain (\%)} \\
\midrule
BC & 4B & 89.17 & 87.16 & \textbf{82.86} & 7.1 \\
BC & 8B & 73.18 & 77.27 & \textbf{68.14} & 6.9 \\
BC & 14B & 147.08 & 167.77 & \textbf{143.15} & 2.7 \\
SWE & 4B & 63.26 & 63.15 & \textbf{59.96} & 5.2 \\
SWE & 8B & 80.12 & 84.13 & \textbf{75.90} & 5.3 \\
SWE & 14B & 95.60 & 107.77 & \textbf{91.90} & 3.9 \\
\bottomrule\end{tabular*}
\end{table}

\begin{figure}[t]
  \centering
  \includegraphics[width=1.0\linewidth]{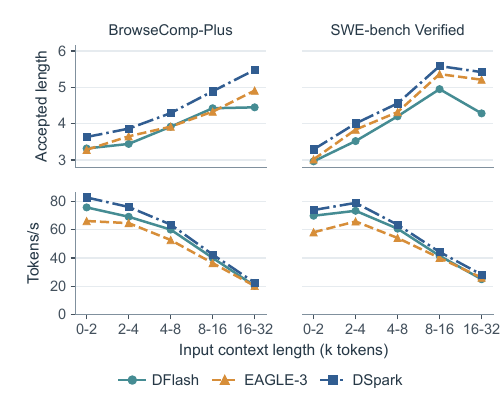}
  \caption{Online baseline acceptance and throughput by context length. Qwen3-8B at $T=0$: upper panels show accepted tokens per round; lower panels show output tokens/s, including prefill.}
  \Description{Four line charts compare DFlash, EAGLE-3, and DSpark on two online workloads. Browsing acceptance rises across the five intervals. Software engineering acceptance peaks at 8k to 16k before falling at 16k to 32k. Throughput is lower in the longest than in the shortest interval in both workloads.}
  \label{fig:diagnosis}
\end{figure}

Across the shortest and longest online context intervals in Figure~\ref{fig:diagnosis}, aggregate acceptance increases while generation throughput decreases. For DSpark with Qwen3-8B at $T=0$, browsing acceptance rises from 3.633 tokens per round below 2k input tokens to 5.479 at 16k to 32k, while throughput falls from 82.73 to 22.54 tokens/s. Software engineering throughput falls from 73.83 to 28.19 tokens/s across the same intervals. This endpoint pattern of higher acceptance and lower throughput holds for all three methods in both workloads. The complete profiles reveal a further distinction: browsing acceptance increases across the five intervals, whereas software engineering acceptance peaks at 8k to 16k and declines in the final interval for every method. Both workloads therefore show lower throughput at long contexts, while the shape of the acceptance curve depends on the workload.

These measurements distinguish draft agreement from overall model-call costs. Accepted length describes agreement on the next continuation; model-call throughput includes the work of processing the available history. At the task level, verification is one component of the complete execution cost:
\begin{equation}
\begin{aligned}
T_{\mathrm{episode}}=\sum_j\bigl(&t_j^{\mathrm{prefill}}+t_j^{\mathrm{draft+verify}}
+t_j^{\mathrm{tool}}+t_j^{\mathrm{other}}\bigr).
\end{aligned}
\label{eq:latency}
\end{equation}
For a fixed continuation length, higher acceptance reduces the number of verification rounds. The contribution of this reduction to task latency depends on the other costs in Equation~\ref{eq:latency}. Across the six greedy online configurations in Table~\ref{tab:latency}, DSpark improves accepted length over DFlash by 10.1\% to 13.8\%, while mean episode latency decreases by 2.7\% to 7.1\%.

The aggregate acceptance curve also combines different decisions. Each context interval contains the tasks and phases reached during execution, weighted by their verification counts. Its trend reflects both agreement within these states and their relative contributions to the workload. A context interval containing a larger share of high-acceptance calls can have a higher aggregate even when agreement falls within a particular phase. This makes the composition of the trajectory an explicit part of the analysis. We therefore examine late actions within matched tasks to resolve how a specific decision type behaves beneath the rising aggregate trend.

\subsubsection{Late Actions within Matched Tasks}
\label{sec:lateactions}
\begin{table}[t]
\caption{Late-action acceptance: Qwen3-8B, $T=0$, step $\geq20$. Tasks are paired across 8k to 16k and 16k to 32k contexts; $\Delta L=L_{16:32}-L_{8:16}$.}
\label{tab:late}
\centering
\fontsize{8}{10}\selectfont
\paperTableSetup
\begin{tabular*}{\linewidth}{@{\extracolsep{\fill}}llccc@{}}
\toprule
\paperTableHead{Workload} & \paperTableHead{Drafter} & \paperTableHead{$L_{8:16}$} & \paperTableHead{$L_{16:32}$} & \paperTableHead{95\% CI of $\Delta L$} \\
\midrule
BC & DFlash & 5.661 & 4.326 & [-1.650, -0.941] \\
BC & EAGLE-3 & 6.007 & 4.897 & [-1.489, -0.612] \\
BC & DSpark & 6.194 & 5.322 & [-1.205, -0.473] \\
SWE & DFlash & 5.333 & 4.762 & [-0.852, -0.257] \\
SWE & EAGLE-3 & 6.149 & 6.304 & [-0.347, 0.716] \\
SWE & DSpark & 6.057 & 6.106 & [-0.238, 0.379] \\
\bottomrule\end{tabular*}
\end{table}

We select action calls at step 20 or later and pair tasks contributing to both the 8k to 16k and 16k to 32k intervals. This comparison holds task identity, execution phase, and the late-stage threshold fixed while examining acceptance at two context ranges. Counters are aggregated within each task and interval; paired task resampling estimates the difference between the resulting acceptance ratios. For Qwen3-8B at $T=0$, DSpark browsing acceptance decreases from 6.194 to 5.322 tokens per round, a difference of $-0.872$ with a pointwise 95\% interval of $[-1.205,-0.473]$. Table~\ref{tab:late} gives corresponding differences of $-1.336$ and $-1.110$ for DFlash and EAGLE-3. The direction is opposite to the aggregate online comparison in Section~\ref{sec:cost}. Resolving the trajectory by phase and progress therefore exposes lower agreement in late browsing decisions even as the overall acceptance curve rises. Appendix~\ref{app:conditional} details the paired estimator.

The corresponding software engineering differences are $-0.571$, $+0.155$, and $+0.049$ for DFlash, EAGLE-3, and DSpark, respectively. DFlash has a negative difference, while the latter two point estimates change little. Under the same phase and progress condition, the workloads exhibit different context profiles. These within-trajectory differences motivate supervision of intermediate assistant decisions, with priorities estimated from agreement in their respective states.

\begin{figure}[t]
  \centering
  \includegraphics[width=0.47\textwidth]{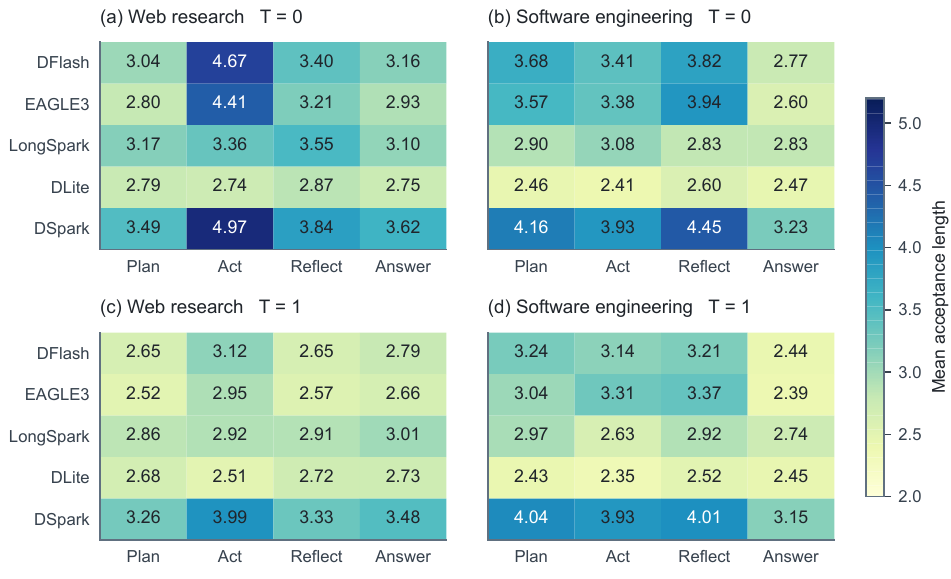}
  \caption{Baseline acceptance by domain and phase. Qwen3-8B results on \corpus{} at $T=0$ and $T=1$, measured in accepted tokens per round. All panels share a common color scale.}
  \Description{Four heatmaps compare five baseline drafters across planning, action, reflection, and answer generation. DSpark has the highest accepted length in every displayed phase. At temperature zero, its highest-acceptance phases are Web research actions and software engineering reflections.}
  \label{fig:phases}
\end{figure}

\subsubsection{Workload and Phase Profiles}
\label{sec:phaseprofiles}
The replay profiles in Figure~\ref{fig:replaycontext} compare methods on identical saved decision inputs. For DSpark with the 8B target at $T=0$, Web research acceptance decreases from 5.242 below 2k tokens to 3.442 at 16k to 32k. Software engineering acceptance increases from 3.073 to 4.321 across the same intervals. These opposing endpoint trends place workload alongside length as an essential condition for interpreting acceptance. The relationship also varies across drafters. At $T=0$, all five baselines have higher acceptance in the longest than in the shortest software engineering interval. In Web research, DFlash and EAGLE-3 share the declining endpoint trend of DSpark, while LongSpark and DLite have higher long-context endpoints. A state-level analysis therefore distinguishes both the workload being served and the draft used to serve it.

Figure~\ref{fig:phases} compares acceptance across phases. For DSpark at $T=0$, Web research actions reach 4.966 tokens per round, compared with 3.488 for planning. Software engineering reflection reaches 4.447, compared with 3.232 for answers. The highest-acceptance phase changes from action to reflection across these domains. Context and phase profiles together describe where the draft aligns well with the target and where additional supervision can be directed.

The practical implication is to estimate training priority at the granularity of an assistant decision, using conditional agreement to distinguish supervision needs among states with comparable history lengths. Intermediate actions and revisions carry information that a trajectory average can conceal, while workload and phase provide the conditions needed to interpret their agreement. Priorities can then be refreshed from the current draft as training proceeds, linking the observed heterogeneity to a concrete allocation problem.

\subsection{Adaptive Draft Supervision}
\label{sec:method}

In Figure~\ref{fig:framework}(c), training couples assistant-turn selection with the share of prediction events assigned to each turn. The allocation policy operates on segments from target-generated trajectories and uses training feedback to revise those shares. The frozen target supplies outputs and features, while the draft retains the DSpark architecture and distillation objective~\cite{dspark}.

\subsubsection{Assistant-Turn Supervision}
Each assistant generation in a target-generated training trajectory becomes a segment $s=(H_s,Y_s)$, comprising its available history and text or tool-call continuation. Training supervises eligible predictions in $Y_s$. Each segment carries state $z_s=(\ell_s,e_s,\omega_s,u_s)$: actual prefix length, execution phase, workload, and preceding tool state, including observations and error or retry events. Tasks are split before expansion. Full-turn expansion supplies intermediate decisions for both fixed and adaptive allocation.

\subsubsection{State-Conditioned Acceptance Gap}
An anchor starts a supervised draft block. For sampled block $b$ with $m_b$ consecutive valid predictions, target and draft distributions $p_{b,k}$ and $q_{b,k}$ on a common vocabulary have overlap
\begin{equation}
\begin{aligned}
a_{b,k}&=\sum_v\min\{p_{b,k}(v),q_{b,k}(v)\}\\
       &=1-\TV(p_{b,k},q_{b,k}).
\end{aligned}
\label{eq:overlap}
\end{equation}
Here $\TV$ denotes total variation distance. This overlap connects distribution agreement to speculative acceptance~\cite{leviathan2023,chen2023}. Over sampled blocks $\mathcal A_s$, we define
\begin{equation}
D_s=1-\frac{\displaystyle\sum_{b\in\mathcal A_s}
                 \sum_{k=1}^{m_b}\prod_{j=1}^{k}a_{b,j}}
                {\displaystyle\sum_{b\in\mathcal A_s}m_b}.
\label{eq:difficulty}
\end{equation}
The prefix products emphasize early disagreement within a block; $D_s\in[0,1]$ measures the resulting normalized gap. The signal uses teacher-forced response prefixes and overlap statistics available in DSpark distillation. Segment feedback is smoothed by trajectory state:
\begin{equation}
\widehat D_s(t)=
\frac{n_s(t)\overline D_s(t)+\kappa\overline D_{g(s)}(t)}
     {n_s(t)+\kappa}.
\label{eq:shrinkage}
\end{equation}
Here $\overline D_s$ is the segment moving average, $n_s$ its decayed visit support, $\overline D_{g(s)}$ the mean of the state group, and $\kappa>0$ the shrinkage strength. Groups combine the attributes in $z_s$, pooling at coarser levels for sparse states. This estimate combines segment feedback with shared state structure as the draft changes.

\subsubsection{Allocation under a Prediction Budget}
An effective prediction event is a valid anchor and prediction-position pair used in an update. Base shares $b_s$, summing to one, distribute supervision across actual prefix-length strata and workloads, with uniform shares among segments in each group. We normalize and clip estimated gaps:
\begin{equation}
r_s(t)=\clip\!\left(
\frac{\widehat D_s(t)+\epsilon}
     {\sum_j b_j\widehat D_j(t)+\epsilon},
r_{\min},r_{\max}\right),
\label{eq:relative}
\end{equation}
Here $\epsilon>0$ stabilizes the ratio and $0<r_{\min}\leq r_{\max}$ bounds relative difficulty. The assigned supervision share is
\begin{equation}
w_s(t)=(1-\lambda)b_s+
\lambda\frac{b_s r_s(t)}{\sum_j b_j r_j(t)}.
\label{eq:allocation}
\end{equation}
With $0\leq\lambda<1$, the allocation retains $w_s\geq(1-\lambda)b_s$; equal estimated gaps recover the base shares. Adaptive exposure follows relative disagreement.

Let $\widehat c_s(t)>0$ estimate the effective events supplied by one segment visit. Event-normalized sampling uses
\begin{equation}
\pi_t(s)=\frac{w_s(t)/\widehat c_s(t)}
                 {\sum_j w_j(t)/\widehat c_j(t)}.
\label{eq:sampling}
\end{equation}
With accurate event-count estimates, this gives expected supervision shares $w_s$. Actual training masks count the effective prediction events consumed toward budget $\mathcal B$. The fixed control uses the same conversion with $w_s=b_s$. The first epoch collects statistics under the base allocation; subsequent epochs refresh the sampling table from training feedback. Development tasks determine scheduling hyperparameters. Scheduling scores are detached, and sampled segments use the original distillation loss. Appendix~\ref{app:allocation} gives the pooling hierarchy, allocation properties, and expected-share derivation.

\section{Experiments}
\label{sec:results}
\begin{table*}[t]
\caption{Phase-level dynamic acceptance at $T=0$ (accepted tokens per round, $L$). Bold marks the highest value per phase and workload. TraceSpec uses adaptive full-turn supervision.}
\label{tab:online}
\centering
\fontsize{8}{10}\selectfont
\paperTableSetup
\setlength{\tabcolsep}{2pt}
\begin{tabular*}{\textwidth}{@{\extracolsep{\fill}}ll*{6}{c}@{\hspace{10pt}}*{6}{c}@{}}
\toprule
\multirow[c]{2}{*}{\paperTableHead{Target}} & \multirow[c]{2}{*}{\paperTableHead{Phase}} & \multicolumn{6}{c}{\textbf{BrowseComp-Plus}} & \multicolumn{6}{c}{\textbf{SWE-bench Verified}} \\
\cmidrule(lr){3-8}\cmidrule(l){9-14}
 &  & \paperTableHead{EAGLE-3} & \paperTableHead{DFlash} & \paperTableHead{LongSpark} & \paperTableHead{DLite} & \paperTableHead{DSpark} & \paperTableHead{TraceSpec} & \paperTableHead{EAGLE-3} & \paperTableHead{DFlash} & \paperTableHead{LongSpark} & \paperTableHead{DLite} & \paperTableHead{DSpark} & \paperTableHead{TraceSpec} \\
\midrule
\multirow{4}{*}{Qwen3-4B} & Plan & 3.674 & 3.581 & 3.112 & 2.804 & 4.031 & \textbf{4.161} & 4.058 & 3.972 & 3.109 & 2.661 & 4.407 & \textbf{4.456} \\
 & Act & 4.605 & 4.072 & 3.260 & 3.281 & 4.617 & \textbf{4.923} & 5.113 & 4.491 & 3.710 & 2.638 & 5.116 & \textbf{5.467} \\
 & Reflect & 4.658 & 4.112 & 3.064 & 3.060 & 4.779 & \textbf{5.123} & 5.118 & 4.624 & 2.956 & 2.665 & 5.258 & \textbf{5.402} \\
 & Answer & 3.910 & 3.463 & 4.186 & 3.261 & 3.980 & \textbf{4.279} & 3.687 & 3.356 & 3.181 & 2.592 & 3.901 & \textbf{4.295} \\
\midrule
\multirow{4}{*}{Qwen3-8B} & Plan & 3.946 & 3.922 & 3.373 & 3.035 & 4.456 & \textbf{4.570} & 3.271 & 3.190 & 2.840 & 2.614 & 3.562 & \textbf{3.656} \\
 & Act & 3.871 & 3.788 & 3.037 & 2.812 & 4.210 & \textbf{4.346} & 4.524 & 4.156 & 3.345 & 2.627 & 4.686 & \textbf{4.935} \\
 & Reflect & 4.315 & 4.156 & 3.103 & 2.926 & 4.688 & \textbf{4.881} & 4.424 & 4.169 & 2.783 & 2.702 & 4.714 & \textbf{4.976} \\
 & Answer & 3.370 & 3.388 & 3.458 & 2.821 & 3.889 & \textbf{3.989} & 3.426 & 3.388 & 3.086 & 2.713 & 3.927 & \textbf{4.217} \\
\midrule
\multirow{4}{*}{Qwen3-14B} & Plan & 3.741 & 3.829 & 3.663 & 3.127 & 4.289 & \textbf{4.486} & 2.887 & 2.746 & 2.780 & 2.473 & 3.123 & \textbf{3.185} \\
 & Act & 4.031 & 4.357 & 3.277 & 2.885 & 4.831 & \textbf{5.018} & 4.064 & 4.146 & 3.301 & 2.542 & 4.616 & \textbf{4.780} \\
 & Reflect & 4.030 & 4.922 & 3.711 & 3.108 & 5.253 & \textbf{5.890} & 3.840 & 3.929 & 3.044 & 2.747 & 4.443 & \textbf{4.499} \\
 & Answer & 3.619 & 4.077 & 3.973 & 2.884 & \textbf{4.535} & 4.512 & 3.232 & 3.504 & 3.375 & 2.798 & 3.975 & \textbf{4.188} \\
\bottomrule\end{tabular*}
\end{table*}

\begin{table*}[t]
\caption{Replay acceptance on \corpus{} ($L$). All methods use identical saved prefixes. Column groups separate temperatures; rows and highlighting follow Table~\ref{tab:online}.}
\label{tab:replay}
\centering
\fontsize{8}{10}\selectfont
\paperTableSetup
\setlength{\tabcolsep}{2pt}
\begin{tabular*}{\textwidth}{@{\extracolsep{\fill}}ll*{6}{c}@{\hspace{10pt}}*{6}{c}@{}}
\toprule
\multirow[c]{2}{*}{\paperTableHead{Target}} & \multirow[c]{2}{*}{\paperTableHead{Phase}} & \multicolumn{6}{c}{\textbf{\corpus{}, $T=0$}} & \multicolumn{6}{c}{\textbf{\corpus{}, $T=1$}} \\
\cmidrule(lr){3-8}\cmidrule(l){9-14}
 &  & \paperTableHead{EAGLE-3} & \paperTableHead{DFlash} & \paperTableHead{LongSpark} & \paperTableHead{DLite} & \paperTableHead{DSpark} & \paperTableHead{TraceSpec} & \paperTableHead{EAGLE-3} & \paperTableHead{DFlash} & \paperTableHead{LongSpark} & \paperTableHead{DLite} & \paperTableHead{DSpark} & \paperTableHead{TraceSpec} \\
\midrule
\multirow{4}{*}{Qwen3-4B} & Plan & 3.899 & 3.585 & 3.611 & 2.618 & 4.161 & \textbf{4.856} & 2.972 & 2.756 & 2.863 & 2.470 & 3.639 & \textbf{4.218} \\
 & Act & 3.704 & 3.419 & 2.788 & 2.549 & 3.953 & \textbf{4.909} & 3.316 & 2.892 & 2.824 & 2.471 & 3.789 & \textbf{4.695} \\
 & Reflect & 3.777 & 3.439 & 3.252 & 2.651 & 4.092 & \textbf{4.859} & 3.303 & 2.810 & 3.004 & 2.499 & 3.752 & \textbf{4.669} \\
 & Answer & 3.089 & 2.803 & 3.013 & 2.578 & 3.294 & \textbf{4.126} & 2.788 & 2.507 & 2.881 & 2.488 & 3.248 & \textbf{3.993} \\
\midrule
\multirow{4}{*}{Qwen3-8B} & Plan & 3.156 & 3.345 & 3.041 & 2.610 & 3.808 & \textbf{4.397} & 2.755 & 2.925 & 2.912 & 2.553 & 3.620 & \textbf{4.060} \\
 & Act & 3.873 & 3.999 & 3.203 & 2.563 & 4.436 & \textbf{5.141} & 3.155 & 3.129 & 2.769 & 2.422 & 3.956 & \textbf{4.630} \\
 & Reflect & 3.504 & 3.578 & 3.221 & 2.751 & 4.095 & \textbf{4.662} & 2.889 & 2.899 & 2.915 & 2.617 & 3.602 & \textbf{4.153} \\
 & Answer & 2.764 & 2.963 & 2.968 & 2.611 & 3.425 & \textbf{4.081} & 2.528 & 2.615 & 2.878 & 2.592 & 3.319 & \textbf{4.086} \\
\midrule
\multirow{4}{*}{Qwen3-14B} & Plan & 3.160 & 3.517 & 3.050 & 2.548 & 3.933 & \textbf{4.584} & 2.793 & 3.080 & 2.901 & 2.485 & 3.654 & \textbf{4.273} \\
 & Act & 3.382 & 3.754 & 3.027 & 2.539 & 4.149 & \textbf{5.053} & 2.914 & 3.385 & 2.845 & 2.418 & 3.936 & \textbf{4.728} \\
 & Reflect & 3.354 & 3.681 & 3.093 & 2.624 & 4.128 & \textbf{4.737} & 2.815 & 3.243 & 2.873 & 2.492 & 3.789 & \textbf{4.247} \\
 & Answer & 2.921 & 3.313 & 3.088 & 2.586 & 3.714 & \textbf{4.282} & 2.600 & 3.028 & 2.941 & 2.537 & 3.536 & \textbf{4.149} \\
\bottomrule\end{tabular*}
\end{table*}

\subsection{Experimental Setup}
\label{sec:setup}
\paragraph{Datasets and evaluation protocols.}
Dynamic evaluation uses 100 tasks from each of BrowseComp-Plus~\cite{browsecomp} and SWE-bench Verified~\cite{swebench}. Browsing agents retrieve and inspect benchmark documents. Repository agents read real repository contents, with simulated feedback for edit and test operations. Fixed-prefix replay evaluates the 2,500 \corpus{} test states described in Section~\ref{sec:corpus}; every method receives the same saved histories and generates one assistant continuation. The training partition supplies separate source tasks for draft adaptation.

\paragraph{Baselines and inference settings.}
We compare TraceSpec with DFlash~\cite{dflash}, EAGLE-3~\cite{eagle3}, DSpark~\cite{dspark}, LongSpark~\cite{longspark}, and DLite~\cite{dlite} using Qwen3-4B, 8B, and 14B targets~\cite{qwen} at $T\in\{0,1\}$, with thinking mode disabled. The three workloads, six methods, three target sizes, and two temperatures give 108 configurations. Baselines use their released target-specific checkpoints. DFlash, DSpark, and TraceSpec use seven-token draft blocks; EAGLE-3 uses the corresponding released configuration. Dynamic episodes allow 50 action steps and 32,768 context tokens; replay allows 36,864 context tokens. Both modes permit up to 4,096 new tokens per call. Acceptance comparisons cover all six methods. Generation-throughput comparisons use the native PyTorch group of DFlash, EAGLE-3, DSpark, DLite, and TraceSpec. LongSpark uses SGLang~\cite{sglang} and contributes to the acceptance comparisons. Episode latency is compared within the original DFlash, EAGLE-3, and DSpark runs. Checkpoint families and additional settings appear in Appendix~\ref{app:protocol}.

\paragraph{Training protocol.}
TraceSpec adapts the corresponding DSpark draft using target-generated trajectories from the training partition. The Qwen3-4B ablations share these trajectories and match effective-event budgets, draft initialization, target model, distillation loss, and optimizer settings. Section~\ref{sec:trainingdesign} specifies the changes to supervision scope, allocation, state estimation, and scoring.

\paragraph{Evaluation metrics.}
For call $i$, let $A_i$ count emitted accepted tokens, including the correction or bonus token, $V_i$ verification rounds, $P_i$ proposed draft tokens, $N_i$ output tokens, and $t_i$ model-call time. We report accepted length $L$, verification yield $R$, and generation throughput $G$:
\begin{equation}
L=\frac{\sum_i A_i}{\sum_i V_i},\quad
R=\frac{\sum_i A_i}{\sum_i(P_i+V_i)},\quad
G=\frac{\sum_i N_i}{\sum_i t_i}.
\label{eq:metrics}
\end{equation}
Model-call time includes prefill and decoding. Episode latency spans the full interactive loop, including tool execution. Ratios use summed counters so that each verification round or generated token contributes to the corresponding aggregate. Pointwise 95\% baseline acceptance intervals use 5,000 bootstrap replicates over original tasks for dynamic execution and source-task clusters for replay, preserving method pairing and recomputing counter ratios.

\begin{table*}[t]
\caption{Qwen3-4B component ablations on accepted length $L$. $A$ and $B$ use fixed allocation; $C_1$ and $C_2$ each modify TraceSpec.}
\label{tab:trainingablation}
\centering
\fontsize{8}{10}\selectfont
\paperTableSetup
\begin{tabular*}{\textwidth}{@{}L{174pt}@{\extracolsep{\fill}\hspace{16pt}}C{42pt}@{\extracolsep{0pt}\hspace{4pt}}C{42pt}@{\extracolsep{\fill}\hspace{16pt}}C{42pt}@{\extracolsep{0pt}\hspace{4pt}}C{42pt}@{\extracolsep{\fill}\hspace{16pt}}C{42pt}@{\extracolsep{0pt}\hspace{4pt}}C{42pt}@{}}
\toprule
\multirow[c]{2}{*}{\paperTableHead{Training variant}} & \multicolumn{2}{c}{\paperTableHead{BrowseComp-Plus}} & \multicolumn{2}{c}{\paperTableHead{SWE-bench Verified}} & \multicolumn{2}{c}{\paperTableHead{TraceSpec-Traj}} \\
\cmidrule(lr){2-3}\cmidrule(lr){4-5}\cmidrule(l){6-7}
 & \paperTableHead{$T=0$} & \paperTableHead{$T=1$} & \paperTableHead{$T=0$} & \paperTableHead{$T=1$} & \paperTableHead{$T=0$} & \paperTableHead{$T=1$} \\
\midrule
\textbf{TraceSpec} & \textbf{4.803} & \textbf{4.145} & \textbf{5.142} & \textbf{4.859} & \textbf{4.753} & \textbf{4.428} \\
\hspace*{1em}w/o intermediate-turn supervision ($A$) & 4.307 & 3.745 & 4.619 & 4.169 & 4.275 & 3.782 \\
\hspace*{1em}w/o adaptive budget allocation ($B$) & 4.473 & 3.904 & 4.849 & 4.417 & 4.602 & 4.169 \\
\hspace*{1em}w/o state-conditioned estimation ($C_1$) & 4.420 & 4.000 & 4.769 & 4.449 & 4.422 & 4.056 \\
\hspace*{1em}w/o prefix-aware scoring ($C_2$) & 4.510 & 4.055 & 4.855 & 4.439 & 4.598 & 4.220 \\
\bottomrule\end{tabular*}
\end{table*}

\begin{table}[t]
\caption{Qwen3-4B inference results. Throughput includes prefill and uses native PyTorch methods. Bold marks column maxima within each metric.}
\label{tab:inference4b}
\centering
\fontsize{8}{10}\selectfont
\paperTableSetup
\begin{tabular*}{\linewidth}{@{}L{42pt}@{\extracolsep{\fill}\hspace{8pt}}C{25pt}@{\extracolsep{0pt}\hspace{3pt}}C{25pt}@{\extracolsep{\fill}\hspace{8pt}}C{25pt}@{\extracolsep{0pt}\hspace{3pt}}C{25pt}@{\extracolsep{\fill}\hspace{8pt}}C{25pt}@{\extracolsep{0pt}\hspace{3pt}}C{25pt}@{}}
\toprule
\multirow[c]{3}{*}{\paperTableHead{Drafter}} & \multicolumn{2}{c}{\paperTableHead{BrowseComp-}} & \multicolumn{2}{c}{\paperTableHead{SWE-bench}} & \multicolumn{2}{c}{\paperTableHead{TraceSpec-}} \\
 & \multicolumn{2}{c}{\paperTableHead{Plus}} & \multicolumn{2}{c}{\paperTableHead{Verified}} & \multicolumn{2}{c}{\paperTableHead{Traj}} \\
\cmidrule(lr){2-3}\cmidrule(lr){4-5}\cmidrule(l){6-7}
 & \paperTableHead{$T=0$} & \paperTableHead{$T=1$} & \paperTableHead{$T=0$} & \paperTableHead{$T=1$} & \paperTableHead{$T=0$} & \paperTableHead{$T=1$} \\
\midrule
\multicolumn{7}{@{}l}{\textit{Verification yield $R$ (\%)}} \\
DFlash & 50.6 & 40.3 & 54.9 & 45.9 & 42.7 & 34.8 \\
EAGLE-3 & 56.9 & 48.1 & 61.3 & 53.0 & 47.8 & 40.3 \\
LongSpark & 40.5 & 38.4 & 43.0 & 37.0 & 40.4 & 36.3 \\
DLite & 40.3 & 32.2 & 33.5 & 31.8 & 32.7 & 31.1 \\
DSpark & 57.7 & \textbf{51.5} & 62.3 & 58.1 & 50.0 & 46.0 \\
TraceSpec & \textbf{59.9} & 51.3 & \textbf{64.2} & \textbf{60.6} & \textbf{61.2} & \textbf{57.1} \\
\midrule
\multicolumn{7}{@{}l}{\textit{Generation throughput $G$ (tokens/s)}} \\
DFlash & 41.75 & 37.37 & 62.86 & 57.99 & 38.93 & 32.06 \\
EAGLE-3 & 42.98 & 44.16 & 61.49 & 57.73 & 39.15 & 33.49 \\
DLite & 31.06 & 38.61 & 39.99 & 39.10 & 31.82 & 30.28 \\
DSpark & 46.20 & 54.19 & 67.94 & 66.42 & 44.25 & 40.28 \\
TraceSpec & \textbf{48.73} & \textbf{56.19} & \textbf{69.66} & \textbf{69.70} & \textbf{53.44} & \textbf{49.37} \\
\bottomrule\end{tabular*}
\end{table}

\subsection{Main Results}
\label{sec:mainresults}

Tables~\ref{tab:online} and~\ref{tab:replay} report phase-specific acceptance: dynamic execution at $T=0$ and replay at both temperatures. Across all settings, TraceSpec leads 68 of 72 comparisons, including all 24 replay and 44 of 48 dynamic settings. On TraceSpec-Traj, gains over DSpark range from 12.1\% to 25.3\% across planning, actions, reflection, and answer generation. Replay compares methods on identical histories, linking the observed gains to stronger draft agreement across the same agent decision states. These improvements span the distinct prediction roles covered by full-turn supervision.

Across the five baselines, DSpark leads overall accepted length in Figure~\ref{fig:acceptanceoverview} and verification yield in all 18 settings. All 72 paired accepted-length comparisons against the other four baselines have positive lower bounds for their pointwise 95\% intervals, supporting DSpark as the base drafter.

Baseline rankings vary with target and workload. EAGLE-3 exceeds DFlash in the 4B and 8B dynamic settings, whereas DFlash leads at 14B. On TraceSpec-Traj, EAGLE-3 is the strongest alternative to DSpark at 4B and DFlash at 8B and 14B.

\subsection{Inference Efficiency}
\label{sec:efficiency}

Table~\ref{tab:inference4b} summarizes Qwen3-4B results. TraceSpec achieves the highest verification yield in five settings and throughput in all six. Relative to DSpark, replay yield increases from 50.0\% to 61.2\% at $T=0$ and from 46.0\% to 57.1\% at $T=1$. Corresponding throughput rises from 44.25 to 53.44 and from 40.28 to 49.37 tokens/s, respectively. These improvements connect greater verification yield to faster generation on shared inputs. Across target sizes, TraceSpec leads throughput in 15 of 18 native PyTorch settings, with replay gains of 12.5\% to 22.6\% over DSpark. Appendix~\ref{app:results} provides the 8B and 14B results and supplementary dynamic comparisons.

DSpark has the lowest mean episode latency in 10 of 12 original three-method settings; DFlash leads at 8B and $T=1$. Table~\ref{tab:latency} reports the greedy results. As Section~\ref{sec:cost} shows, episode latency combines verification, history processing, and tool costs, distinguishing task completion time from generation throughput.

\subsection{Effect of Sampling Temperature}
\label{sec:temperature}
Figure~\ref{fig:acceptanceoverview} shows lower accepted length at $T=1$ in all 45 matched baseline comparisons. On TraceSpec-Traj, DSpark decreases by 6.5\% to 8.7\%, compared with 11.0\% to 18.2\% for DFlash and 13.5\% to 15.4\% for EAGLE-3. TraceSpec leads all 12 replay phase comparisons at $T=1$ and exceeds each ablated Qwen3-4B variant at both temperatures in Table~\ref{tab:trainingablation}.

For Qwen3-4B on BrowseComp-Plus, DSpark throughput increases from 46.20 to 54.19 tokens/s at $T=1$ even as acceptance falls. On fixed-prefix replay, throughput decreases from 44.25 to 40.28 tokens/s. Online timing reflects the workload generated at each temperature, while replay holds the input history constant.

\subsection{Training Ablations}
\label{sec:trainingdesign}

Table~\ref{tab:trainingablation} compares TraceSpec with four variants on shared target-generated trajectories under the same effective-event budget. $A$ supervises the final assistant response with fixed allocation; $B$ adds intermediate turns while retaining fixed allocation. Their comparison isolates supervision scope, and TraceSpec versus $B$ isolates adaptive allocation. The remaining variants each modify the full method: $C_1$ replaces the state-group prior in Equation~\ref{eq:shrinkage} with a global prior while preserving segment statistics; $C_2$ replaces the prefix products in Equation~\ref{eq:difficulty} with a mean of prediction gaps.

Full-turn supervision improves accepted length by 3.9\% to 10.2\% over final-turn training across the six workload and temperature settings. Adaptive allocation adds 3.3\% to 10.0\% over fixed full-turn supervision, with the largest increment on SWE-bench Verified at $T=1$, from 4.417 to 4.859. These comparisons show that intermediate responses supply useful training signals and that their allocation contributes beyond broader supervision coverage.

TraceSpec exceeds $C_1$ by 3.6\% to 9.2\% and $C_2$ by 2.2\% to 9.5\% across all six settings. On TraceSpec-Traj at $T=1$, accepted length rises from 4.056 with global-prior estimation and 4.220 with position-wise scoring to 4.428 with the complete method. The former comparison supports estimating difficulty within trajectory conditions; the latter supports accounting for the dependence of later acceptance on earlier predictions. Together, they connect the allocation gains to the state heterogeneity and sequential acceptance structure motivating the method. Appendix~\ref{app:allocation} specifies the controls.
\section{Conclusion}
TraceSpec connects trajectory-resolved evaluation with adaptive draft supervision for LLM agents. TraceSpec-Traj preserves intermediate decision states for controlled replay, complementing dynamic execution. Trajectory analysis shows that drafting difficulty varies with workload, phase, and task progress, including a reversal between aggregate browsing trends and matched late-action behavior. These findings guide full-turn supervision and adaptive allocation through state-conditioned, prefix-aware acceptance gaps. Controlled ablations identify gains from supervision coverage, allocation, and scoring; TraceSpec improves replay generation throughput by 12.5\% to 22.6\% over DSpark. The study establishes an empirical basis for allocating limited draft supervision across heterogeneous agent decision states.

\bibliographystyle{ACM-Reference-Format}
\bibliography{references}
\appendix
\section{Model Configuration and Execution Protocol}
\label{app:protocol}
All runs use the target-specific Qwen3 tokenizer and non-thinking chat template for prefix measurement and generation. DFlash, EAGLE-3, and DSpark load the released DeepSpec checkpoint families matched to each target size, with seven-token draft blocks or the corresponding EAGLE-3 configuration. Confidence-based early stopping is disabled for DSpark. LongSpark uses the target-specific Hehy checkpoints~\cite{longspark}, and DLite uses the SJTU-DAI-Sys checkpoints~\cite{dlite}. Section~\ref{sec:setup} specifies generation limits and timing scopes. Dynamic task selection uses seed 980406; reflection occurs every five steps. Each replay invocation retains the saved history and tool observations, using the same tokenization conventions as dynamic execution.

\section{Corpus Provenance and Interaction Coverage}
\label{app:corpus}
\begin{table}[h]
\caption{Source composition of the constructed trajectory corpus.}
\label{tab:sources}
\centering
\fontsize{8}{10}\selectfont
\paperTableSetup
\begin{tabular*}{\linewidth}{@{}l@{\extracolsep{\fill}\hspace{12pt}}R{38pt}@{\extracolsep{0pt}\hspace{12pt}}R{38pt}@{}}
\toprule
\paperTableHead{Source} & \paperTableHead{Training} & \paperTableHead{Test} \\
\midrule
DeepResearch-traj~\cite{ipfdata} & 5,000 & 1,250 \\
Nebius SWE-agent~\cite{nebiusdata} & 3,806 & 972 \\
R2E-Gym SFT~\cite{r2edata} & 1,031 & 243 \\
R2E-Gym testing agent~\cite{r2etestdata} & 163 & 35 \\
\bottomrule\end{tabular*}
\end{table}

Table~\ref{tab:sources} summarizes the 5,795 training and 1,380 test source tasks. Nominal context strata of 2k, 4k, 8k, 16k, and 32k contribute 18\%, 20\%, 23\%, 21\%, and 18\% of each split; evaluation and state assignment use actual tokenized lengths. Both splits cover all 40 domain, context, and phase combinations. Median test-history message counts are 30 for Web research and 20 for software engineering; median tool-message counts are nine in both domains. Tool messages contribute 80.4\% and 57.1\% of aggregate test-prefix characters, respectively. Decision histories enable analysis of how similar context lengths accompany different prediction requirements. Joint workload and phase coverage permits these conditions to be examined within shared context strata, while saved tool observations preserve the evidence available when the next assistant continuation is generated. Corpus quotas select decision states; assistant-turn expansion defines supervision candidates while evaluation retains the curated test distribution.

\paragraph{Task isolation.}
Collection-qualified query identifiers define browsing tasks; hashes of problem text after scaffold removal identify software tasks. Deterministic task hashing assigns an 80\%/20\% training/test split before assistant-turn expansion, keeping each task's rollouts together. Source-task intersection is zero. Qwen3-4B token 3-gram checks, normalized by the test-question 3-gram count, find zero training/test pairs with overlap at least 0.5. BrowseComp-Plus query identifiers and corpus task identifiers use distinct namespaces; question-text comparisons therefore assess overlap independently of numerical identifiers. Cross-checking all 824 BrowseComp-Plus queries against 3,769 browsing tasks from both partitions yields a maximum query-normalized overlap 0.1818, below the 0.2 threshold. All 500 SWE-bench Verified instances are excluded before splitting through task-key and original-label matching. The training partition supplies adaptation and scheduling feedback; development tasks reserved within this split before segment generation guide hyperparameter selection. Replay and dynamic tasks serve only post-training evaluation.

\section{Supplementary Results and Paired Estimation}
\label{app:results}
\begin{table*}[t]
\caption{Dynamic acceptance at $T=1$. Metric and highlighting follow Table~\ref{tab:online}.}
\label{tab:onlinesampling}
\centering
\fontsize{8}{10}\selectfont
\paperTableSetup
\setlength{\tabcolsep}{2pt}
\begin{tabular*}{\textwidth}{@{\extracolsep{\fill}}ll*{6}{c}@{\hspace{10pt}}*{6}{c}@{}}
\toprule
\multirow[c]{2}{*}{\paperTableHead{Target}} & \multirow[c]{2}{*}{\paperTableHead{Phase}} & \multicolumn{6}{c}{\textbf{BrowseComp-Plus}} & \multicolumn{6}{c}{\textbf{SWE-bench Verified}} \\
\cmidrule(lr){3-8}\cmidrule(l){9-14}
 &  & \paperTableHead{EAGLE-3} & \paperTableHead{DFlash} & \paperTableHead{LongSpark} & \paperTableHead{DLite} & \paperTableHead{DSpark} & \paperTableHead{TraceSpec} & \paperTableHead{EAGLE-3} & \paperTableHead{DFlash} & \paperTableHead{LongSpark} & \paperTableHead{DLite} & \paperTableHead{DSpark} & \paperTableHead{TraceSpec} \\
\midrule
\multirow{4}{*}{Qwen3-4B} & Plan & 3.168 & 2.813 & 3.072 & 2.528 & 3.471 & \textbf{3.614} & 3.454 & 3.247 & 2.800 & 2.459 & 3.934 & \textbf{3.987} \\
 & Act & 3.878 & 3.263 & 3.074 & 2.554 & 4.146 & \textbf{4.228} & 4.423 & 3.800 & 3.069 & 2.510 & 4.841 & \textbf{5.246} \\
 & Reflect & 4.032 & 3.194 & 2.894 & 2.532 & 4.160 & \textbf{4.361} & 4.592 & 3.847 & 2.664 & 2.540 & 4.845 & \textbf{5.054} \\
 & Answer & 3.620 & 3.172 & 3.736 & 2.561 & 4.200 & \textbf{4.289} & 2.949 & 2.647 & 3.034 & 2.505 & 3.426 & \textbf{3.713} \\
\midrule
\multirow{4}{*}{Qwen3-8B} & Plan & 3.036 & 3.021 & 3.175 & 2.790 & \textbf{3.639} & 3.583 & 2.495 & 2.556 & 2.539 & 2.397 & 3.016 & \textbf{3.044} \\
 & Act & 2.965 & 2.895 & 2.813 & 2.574 & 3.481 & \textbf{3.613} & 3.379 & 3.038 & 2.953 & 2.465 & 3.891 & \textbf{4.119} \\
 & Reflect & 3.064 & 2.916 & 2.820 & 2.576 & 3.588 & \textbf{3.745} & 3.150 & 2.949 & 2.592 & 2.465 & 3.713 & \textbf{3.974} \\
 & Answer & 3.141 & 3.115 & 3.380 & 2.719 & 3.792 & \textbf{3.861} & 2.733 & 2.762 & 2.938 & 2.631 & 3.470 & \textbf{3.696} \\
\midrule
\multirow{4}{*}{Qwen3-14B} & Plan & 2.975 & 2.936 & 3.236 & 2.758 & \textbf{3.580} & 3.484 & 2.675 & 2.563 & 2.712 & 2.412 & 2.952 & \textbf{3.091} \\
 & Act & 2.957 & 3.198 & 2.950 & 2.599 & 3.712 & \textbf{3.859} & 3.406 & 3.502 & 2.993 & 2.448 & 3.966 & \textbf{4.404} \\
 & Reflect & 3.072 & 3.372 & 3.213 & 2.785 & 3.969 & \textbf{4.199} & 3.011 & 3.097 & 2.852 & 2.557 & 3.592 & \textbf{3.902} \\
 & Answer & 3.336 & 3.786 & 3.886 & 2.814 & \textbf{4.637} & 4.519 & 2.797 & 3.099 & 3.175 & 2.704 & 3.753 & \textbf{3.804} \\
\bottomrule\end{tabular*}
\end{table*}

\begin{table*}[t]
\caption{Verification yield $R$ (\%) for Qwen3-8B and 14B, extending Table~\ref{tab:inference4b}. Bold marks column maxima within each target.}
\label{tab:yield}
\centering
\fontsize{8}{10}\selectfont
\paperTableSetup
\begin{tabular}{@{}L{44pt}@{\hspace{12pt}}L{42pt}@{\hspace{16pt}}C{38pt}@{\hspace{4pt}}C{38pt}@{\hspace{16pt}}C{38pt}@{\hspace{4pt}}C{38pt}@{\hspace{16pt}}C{38pt}@{\hspace{4pt}}C{38pt}@{}}
\toprule
\multirow[c]{2}{*}{\paperTableHead{Target}} & \multirow[c]{2}{*}{\paperTableHead{Drafter}} & \multicolumn{2}{c}{\paperTableHead{BrowseComp-Plus}} & \multicolumn{2}{c}{\paperTableHead{SWE-bench Verified}} & \multicolumn{2}{c}{\paperTableHead{TraceSpec-Traj}} \\
\cmidrule(lr){3-4}\cmidrule(lr){5-6}\cmidrule(l){7-8}
 &  & \paperTableHead{$T=0$} & \paperTableHead{$T=1$} & \paperTableHead{$T=0$} & \paperTableHead{$T=1$} & \paperTableHead{$T=0$} & \paperTableHead{$T=1$} \\
\midrule
\multirow{6}{*}{Qwen3-8B} & DFlash & 48.9 & 37.1 & 50.4 & 37.2 & 44.1 & 36.6 \\
 & EAGLE-3 & 49.9 & 38.1 & 54.0 & 40.2 & 42.5 & 35.9 \\
 & LongSpark & 39.2 & 36.5 & 39.6 & 35.5 & 39.2 & 35.9 \\
 & DLite & 36.3 & 33.1 & 33.3 & 31.1 & 33.2 & 32.0 \\
 & DSpark & 54.9 & \textbf{45.0} & 56.9 & 47.2 & 50.3 & 45.9 \\
 & TraceSpec & \textbf{55.3} & 44.7 & \textbf{58.6} & \textbf{48.3} & \textbf{59.4} & \textbf{54.3} \\
\midrule
\multirow{6}{*}{Qwen3-14B} & DFlash & 55.8 & 40.7 & 49.9 & 41.8 & 45.2 & 40.2 \\
 & EAGLE-3 & 50.4 & 38.1 & 49.0 & 40.7 & 40.8 & 35.2 \\
 & LongSpark & 43.2 & 38.6 & 40.1 & 36.9 & 38.3 & 36.1 \\
 & DLite & 37.6 & 33.9 & 33.1 & 31.6 & 32.4 & 31.2 \\
 & DSpark & 61.5 & \textbf{48.1} & 55.9 & 47.9 & 50.5 & 47.2 \\
 & TraceSpec & \textbf{63.4} & 47.7 & \textbf{56.2} & \textbf{50.9} & \textbf{60.4} & \textbf{56.1} \\
\bottomrule\end{tabular}
\end{table*}

\begin{table*}[t]
\caption{Generation throughput (output tokens/s) for Qwen3-8B and 14B. Timing scope and highlighting follow Table~\ref{tab:inference4b}.}
\label{tab:throughput}
\centering
\fontsize{8}{10}\selectfont
\paperTableSetup
\begin{tabular}{@{}L{44pt}@{\hspace{12pt}}L{42pt}@{\hspace{16pt}}C{38pt}@{\hspace{4pt}}C{38pt}@{\hspace{16pt}}C{38pt}@{\hspace{4pt}}C{38pt}@{\hspace{16pt}}C{38pt}@{\hspace{4pt}}C{38pt}@{}}
\toprule
\multirow[c]{2}{*}{\paperTableHead{Target}} & \multirow[c]{2}{*}{\paperTableHead{Drafter}} & \multicolumn{2}{c}{\paperTableHead{BrowseComp-Plus}} & \multicolumn{2}{c}{\paperTableHead{SWE-bench Verified}} & \multicolumn{2}{c}{\paperTableHead{TraceSpec-Traj}} \\
\cmidrule(lr){3-4}\cmidrule(lr){5-6}\cmidrule(l){7-8}
 &  & \paperTableHead{$T=0$} & \paperTableHead{$T=1$} & \paperTableHead{$T=0$} & \paperTableHead{$T=1$} & \paperTableHead{$T=0$} & \paperTableHead{$T=1$} \\
\midrule
\multirow{5}{*}{Qwen3-8B} & DFlash & 42.96 & 51.35 & 48.95 & 49.18 & 37.02 & 31.07 \\
 & EAGLE-3 & 40.33 & 41.20 & 45.98 & 41.32 & 32.68 & 27.81 \\
 & DLite & 34.51 & 44.86 & 37.19 & 35.99 & 30.69 & 28.45 \\
 & DSpark & 46.79 & \textbf{54.94} & 52.63 & 51.93 & 40.44 & 37.06 \\
 & TraceSpec & \textbf{48.25} & 53.92 & \textbf{54.30} & \textbf{55.12} & \textbf{47.14} & \textbf{41.70} \\
\midrule
\multirow{5}{*}{Qwen3-14B} & DFlash & 24.41 & 25.80 & 33.86 & 30.61 & 25.81 & 23.72 \\
 & EAGLE-3 & 20.73 & 22.22 & 29.49 & 28.12 & 21.77 & 19.37 \\
 & DLite & 19.73 & 24.29 & 27.04 & 25.76 & 20.58 & 20.00 \\
 & DSpark & 25.25 & \textbf{30.26} & 35.58 & \textbf{35.73} & 27.36 & 25.74 \\
 & TraceSpec & \textbf{26.60} & 29.67 & \textbf{36.09} & 35.14 & \textbf{31.12} & \textbf{29.60} \\
\bottomrule\end{tabular}
\end{table*}

Table~\ref{tab:onlinesampling} extends the dynamic comparisons in Section~\ref{sec:results} to $T=1$. Tables~\ref{tab:yield} and~\ref{tab:throughput} extend the Qwen3-4B results in Table~\ref{tab:inference4b} to 8B and 14B, retaining the same metrics and runtime grouping. Across all target sizes and temperatures, TraceSpec increases replay verification yield over DSpark by 8.4 to 11.2 percentage points. Greedy dynamic throughput improves by 1.4\% to 5.5\% across the six workload and target-size settings. Among the released baselines, DSpark replay acceptance exceeds the strongest alternative by 7.3\% to 25.1\%; the largest gain occurs at 8B and $T=1$ against DFlash, with a pointwise 95\% interval of 21.9\% to 28.3\%.

Temperature also changes the workload encountered during dynamic execution. On BrowseComp-Plus with the 14B target, increasing $T$ from zero to one lowers DSpark acceptance from 4.837 to 3.796 while throughput rises from 25.25 to 30.26 tokens/s. Mean action steps fall from 21.73 to 15.72, mean input length from 13,035 to 9,245 tokens, and episode latency from 143.15 to 91.08 seconds. With fixed replay prefixes, DSpark throughput decreases by 5.9\% to 9.0\% across target sizes. The replay acceptance reductions for LongSpark and DLite are 5.9\% to 10.3\% and 3.6\% to 4.8\%, respectively.

Paired replay intervals compare DSpark with each alternative at matched target sizes and temperatures. Estimation uses 5,000 paired source-task bootstrap replicates with seed 20261021. Decisions from the same source task can span several context strata and execution phases while sharing earlier assistant messages and tool observations. Source-task resampling carries the same multiplicity to every retained decision from a task and to every drafter. This alignment preserves both dependence across decisions and covariance between methods. Recomputing the pooled ratios within each replicate keeps the interval estimator consistent with the verification-round weighting of the reported means.

\section{Estimation for Matched Late Actions}
\label{app:conditional}
The estimates in Table~\ref{tab:late} follow the matched-task analysis in Section~\ref{sec:lateactions}, using late-action calls from both context intervals. Within each eligible task, emission and verification counters are summed separately for the two intervals. Workload, target size, temperature, phase, and the late-stage threshold remain fixed. Bootstrap replicates resample tasks while carrying both interval totals for every selected task. Each replicate first recomputes the two pooled acceptance ratios and then takes their difference. This preserves the dependence between observations from the same task and weights each interval by its verification rounds. The resulting interval estimates the contrast among tasks represented in both conditions.

\section{Adaptive Scheduling and Training Controls}
\label{app:allocation}
\paragraph{Training segments and state estimation.}
Each tool observation remains associated with its executed call. Segments preserve actual prefix lengths and preceding tool events, and repeated decision positions are consolidated. State estimates pool statistics over length, phase, workload, and tool state, then over length, phase, and workload; length and workload; and finally the global pool. The number of distinct training trajectories determines which pooling level has adequate support. In Equation~\ref{eq:shrinkage}, $n_s$ measures decayed visits to segment $s$ and controls the weight placed on its own estimate. Thus, trajectory counts determine pooling granularity, while visit counts determine shrinkage. Segment means enter group statistics with equal weight, and unvisited segments inherit the group mean. State-conditioned estimates distinguish disagreement associated with different decision roles within the same length stratum. Their updates follow the changing draft, so supervision priorities can shift as some segments become better aligned. The common effective-event budget makes these changes interpretable as a redistribution of supervised predictions across the trajectory.

\paragraph{Epoch-wise allocation updates.}
Allocation remains fixed within each epoch. The first epoch uses base shares $b_s$ and records acceptance gaps and effective prediction counts. At each epoch boundary, this feedback updates segment and group estimates, supervision shares, and event-normalized sampling probabilities through Equations~\ref{eq:shrinkage} to~\ref{eq:sampling}. The next epoch uses these probabilities with the unchanged distillation objective.

\paragraph{Coverage and allocation bounds.}
Equation~\ref{eq:allocation} combines two normalized distributions, giving $\sum_s w_s=1$. Its nonnegative adaptive component ensures $w_s\geq(1-\lambda)b_s$ for every candidate segment. Equal smoothed gaps yield equal clipped scores, recovering the base allocation $b_s$. Clipping limits the ratio between adaptive multipliers, and the base component provides a lower bound on exposure for each represented state.

\paragraph{Expected supervision shares.}
Let $c_s$ be the expected effective-event count for a visit. When the estimate equals $c_s$, the expected event share is
\begin{equation}
\frac{\pi_t(s)c_s}{\sum_j\pi_t(j)c_j}=w_s(t).
\label{eq:eventshare}
\end{equation}
Writing $Z_t=\sum_j w_j(t)/c_j$, Equation~\ref{eq:sampling} gives $\pi_t(s)c_s=w_s(t)/Z_t$. Summing across segments gives $1/Z_t$ and establishes Equation~\ref{eq:eventshare}. Hence $w_s$ specifies the expected share of supervised predictions, whereas $\pi_t(s)$ determines how often a segment is visited. Segments supplying more predictions per visit need fewer visits to attain the same share. Actual training masks determine cumulative budget consumption, including the complete final update; reusing a prediction in a later visit counts as another event. The fixed and adaptive conditions use the same event accounting, sampling conversion, and loss normalization, with $w_s$ controlling the difference in training exposure.

\paragraph{Qwen3-4B training and resource use.}
The complete Qwen3-4B run adapts DSpark block-7 for three epochs and 9,361 optimizer updates, with packed sequences of 32,768 tokens and a fixed 512-anchor cap. Fused AdamW uses a learning rate $6\times10^{-4}$, zero weight decay, cosine decay, and a 0.04 warmup fraction. Distillation weights are 0.1 for cross entropy and 1.8 for total variation, with confidence-head coefficient 1.0 and gradient norm clipping at 1.0. Master weights use FP32; hidden states and autocast use BF16. Data sampling and training seeds are 20260901 and 42; shrinkage strength is $\kappa=8$. One A800 80GB GPU trains the draft and a second serves target hidden states using vLLM 0.18.0. This run takes 9\,h 42\,min 26\,s for training, validation, and on-demand target service, corresponding to 19.41 GPU hours of wall-clock allocation.

\paragraph{Scheduler settings and ablation controls.}
The scheduler sets $\lambda=0.5$, clips relative gaps to $[0.5,2]$, and updates allocation once per epoch. Base quotas are equal across the five represented length strata, with equal workload shares. Smoothing and support parameters are selected on the development partition and fixed before evaluation. $B$ and TraceSpec share the segment population, event accounting, and distillation objective; the allocation changes from $b_s$ to $w_s(t)$.

In $C_1$, the global mean replaces $\overline D_{g(s)}$ while the segment mean, visit support, and shrinkage weight remain unchanged. In $C_2$, individual gaps $1-a_{b,k}$ are averaged over valid prediction positions; subsequent smoothing, clipping, and sampling follow the full method. In Equation~\ref{eq:difficulty}, an early overlap contributes to several prefix products, so early disagreement reduces estimated agreement across the remaining positions within the block. A position-wise average assigns separate contributions to those gaps. The $C_2$ comparison therefore examines whether retaining this ordering improves the distribution of supervision across candidate segments. Inference settings, hardware, and concurrency are fixed across ablations. Replay supplies identical histories, aligning their acceptance measurements at common decision boundaries.

\end{document}